\documentclass[10pt,twocolumn,letterpaper]{article}

\usepackage[pagenumbers]{cvpr} % To force page numbers, e.g. for an arXiv version

\usepackage[dvipsnames]{xcolor}

\usepackage{multirow}

\definecolor{cvprblue}{rgb}{0.21,0.49,0.74}
\usepackage[pagebackref,breaklinks,colorlinks,citecolor=cvprblue]{hyperref}

\def\paperID{*****} % *** Enter the Paper ID here
\def\confName{CVPR}
\def\confYear{2024}

\title{Boosting Continuous Emotion Recognition with Self-Pretraining using Masked Autoencoders, Temporal Convolutional Networks, and Transformers}

\author{Weiwei Zhou, Jiada Lu, Chenkun Ling, Weifeng Wang, Shaowei Liu\\
Chinatelecom Cloud\\
{\tt \small \{zhouweiwei,lujiada,lingchengk,wangweifeng,liusw12\}@chinatelecom.cn}
}

\begin{document}
\maketitle
\begin{abstract}
Human emotion recognition holds a pivotal role in facilitating seamless human-computer interaction. This paper delineates our methodology in tackling the Valence-Arousal (VA) Estimation Challenge, Expression (Expr) Classification Challenge, and Action Unit (AU) Detection Challenge within the ambit of the 6th Workshop and Competition on Affective Behavior Analysis in-the-wild (ABAW). Our study advocates a novel approach aimed at refining continuous emotion recognition. We achieve this by initially harnessing pre-training with Masked Autoencoders (MAE) on facial datasets, followed by fine-tuning on the aff-wild2 dataset annotated with expression (Expr) labels. The pre-trained model serves as an adept visual feature extractor, thereby enhancing the model's robustness. Furthermore, we bolster the performance of continuous emotion recognition by integrating Temporal Convolutional Network (TCN) modules and Transformer Encoder modules into our framework.
\end{abstract}

\section{Introduction}
\label{sec:intro}

Facial Expression Recognition (FER) holds immense potential across a spectrum of applications, ranging from discerning emotions in videos to bolstering security through facial recognition systems, and even enriching virtual reality experiences. While significant strides have been made in various facial-related tasks, such as face and attribute recognition, the nuanced realm of emotional comprehension remains a challenge. 

The intricacies of emotional expressions often present subtle differentiations that can introduce ambiguity or uncertainty in accurately perceiving emotions. Consequently, this complexity poses hurdles in effectively assessing an individual's emotional state. One of the primary obstacles lies in the inadequacy of existing FER datasets to encapsulate the breadth and depth of human emotional expressions, hindering the development of robust models. Efforts to expand and diversify these datasets are imperative to enhance the efficacy and reliability of FER systems.

The appearance of AffWild  and AffWild2 dataset and the corresponding challenges \cite{kollias2023abaw2, kollias2023multi, kollias2019expression, kollias2022abaw, kollias2021analysing, kollias2020analysing, kollias2021distribution, kollias2021affect, kollias2019face, kollias2019deep, zafeiriou2017aff, 2303.01498}
boost the development of affective recognition study. The Aff-Wild2 dataset contains about 600 videos with around 3M frames. The dataset is annotated with three different affect attributes: a) dimensional affect with valence and arousal; b) six basic categorical affect; c) action units of facial muscles. To facilitate the utilization of the Aff-Wild2 dataset, the 6th 
 ABAW\cite{kollias20246th} competition was organized for affective behavior analysis in the wild.

Due to the significant success achieved by pre-training models like MAE. In the past, we attempt to utilize the MAE pre-training method as a visual feature extractor on a facial expression dataset. Subsequently, we employ TCN and Transformer for continuous emotion recognition. Our approach results in a significant improvement in the evaluation accuracy of Valence-Arousal Estimation, Action Unit Detection, and Expression Classification.

The remaining parts of the paper are presented as follows: Sec \ref{sec:RelatedWork} describe the study of facial emotion recognition. Sec \ref{sec:method} describes our methodology; Sec \ref{sec:experiment} describes the experiment details and the result; Sec \ref{sec:conclusion} is the conclusion of the paper.

\section{Related Work}
\label{sec:RelatedWork}

Previous studies have proposed some useful networks on the Aff-wild2 dataset. Kuhnke et al. \cite{kuhnke2020two} combined vision and audio information in the video and constructed a two-stream network for emotion recognition, achieving high performance. Yue Jin et al. \cite{jin2021multi} proposed a transformer-based model to merge audio and visual features.

NetEase \cite{zhang2023multi} utilized the visual information from a Masked Autoencoder (MAE) model that had been pre-trained on a large-scale face image dataset in a self-supervised manner. Next, the MAE encoder was fine-tuned on the image frames from the Aff-wild2 for AU, EXPR, and VA tasks, which could be regarded as static and uni-modal training. Additionally, multi-modal and temporal information from the videos were leveraged, and a transformer-based framework was implemented to fuse the multi-modal features.

SituTech \cite{liu2023multi} utilized multi-modal feature combinations extracted by several different pre-trained models, which were applied to capture more effective emotional information.

Temporal Convolutional Network (TCN) was proposed by Colin Lea et al. \cite{lea2016temporal}, which hierarchically captured relationships at low-, intermediate-, and high-level time scales. Jin Fan et al. \cite{fan2021parallel} proposed a model with a spatial-temporal attention mechanism to catch dynamic internal correlations with stacked TCN backbones to extract features from different window sizes.

The Transformer mechanism proposed by Vaswani et al. \cite{vaswani2017attention} has achieved high performance in many tasks, so many researchers exploited the Transformer for affective behavior studies. Zhao et al. \cite{zhao2021former} proposed a model with spatial and temporal Transformers for facial expression analysis. Jacob et al. \cite{9577264} proposed a network to learn the relationship between action units with a transformer correlation module.

Inspired by the previous work, in this paper, we propose to use MAE as a feature extractor and design a model consisting of TCN and Transformer to enhance the performance of emotion recognition.

\section{Methodology}
\label{sec:method}

In this section, we describe in detail our proposed method for tackling the three challenging tasks of affective behavior analysis in the wild that are addressed by the 6th ABAW Competition: Valence-Arousal Estimation, Expr Classification, and AU Detection. We explain how we design our model architecture, data processing, and training strategy for each task.

\subsection{MAE Pre-training}

Inspired by Netease, we conduct pre-training of our MAE on a facial image dataset. To this end, we also curate a large-scale dataset of facial expressions to learn facial features, consisting of AffectNet, RAF-DB, FER2013, and FER+. Subsequently, the MAE model is pre-trained on this dataset in a self-supervised manner. Specifically, our MAE consists of a ViT-Base encoder and a ViT decoder. The pre-training process of MAE follows a masked-reconstruction method, where images are first divided into a series of patches (16x16), with 75\% of these patches randomly masked. These masked images are then fed into the MAE encoder, while the MAE decoder is tasked with reconstructing the complete image. The loss function for MAE pre-training is pixel-level L2 loss, aiming to minimize the difference between the reconstructed image and the target image. Once self-supervised learning is completed, the MAE decoder is removed and replaced with fully connected layers connected to the MAE encoder. Subsequently, Expr labels are fine-tuned to obtain a feature extractor more aligned with the distribution of aff-wild2 data.

\subsection{Temporal Convolutional Network}

Videos are first split into segments with a window size $w$ and stride $s$. Given the segment window $w$ and stride $s$, a video with $n$ frames would be split into $[n/s] + 1$ segments, where the $i$-th segment contains frames$\left\{F_{(i-1) *s+1}, \ldots, F_{(i-1) * s+w}\right\}$.

In other words, videos are cut into some overlapping chunks, each with a fixed number of frames. The purpose of doing this is to break down the video into smaller parts that are easier to process and analyze. Each chunk has some degree of overlap with the previous and next ones so that no information in the video is missed.

We denote visual features as  $f_i$ corresponding to the $i$-th segment extracted by pre-trained and fine-tuned ViT-Base encoder.

Visual feature is fed into a dedicated Temporal Convolutional Network (TCN) for temporal encoding, which can be formulated as follows: 
$$ g_i=\text { TCN }\left(f_i\right) $$

This means that we use a special type of neural network that can capture the temporal patterns and dependencies of the features over time. The TCN takes the input feature vector and applies a series of convolutional layers with different kernel sizes and dilation rates to produce an output feature vector. The output feature vector has the same length as the input feature vector but contains more information about the temporal context. For example, the TCN can learn how the image changes over time in each segment of the video. 

\subsection{Temporal Encoder}
We utilize a transformer encoder to model the temporal information in the video segment as well, which can be formulated as follows: 
$$ h_i=\text { TransformerEncoder }\left(g_i\right). $$ 
The Transformer encoder only models the context within a single segment, thereby ignoring the dependencies between frames across segments. To account for the context of different frames, overlapping between consecutive segments can be employed, thus enabling the capture of the dependencies between frames across segments, which means $s \leq w$.

We use another type of neural network that can learn the relationships and interactions among the features within each segment. The transformer encoder takes the input feature vector and applies a series of self-attention layers and feed-forward layers to produce an output feature vector. The output feature vector has more semantic meaning and representation power than the input feature vector. For example, the transformer encoder can learn how different parts of the image relate to each other in each segment of the video. However, the transformer encoder does not consider how different segments of the video are connected or influenced by each other. To solve this problem, we can make some segments overlap with each other so that some frames are shared by two or more segments. This way, we can capture some information about how different segments affect each other. The degree of overlap is controlled by two parameters: $s$ is the length of a segment and $w$ is the sliding window size. If $s$ is smaller than or equal to $w$, then there will be some overlap between consecutive segments.

\subsubsection{Prediction}
After the temporal encoder, the features $h_i$ are finally fed into MLP for regression, which can be formulated as follows: 
$$ y_i= \text{MLP} (h_i) $$
where $y_i $ are the predictions of $i$-th segment. For VA challenge, $y_i \in \mathbb{R}^{l \times 2}$. For Expr challenge, $y_i \in \mathbb{R}^{l \times 8}$. For AU challenge, $y_i \in \mathbb{R}^{l \times 12}$ .

The prediction vector contains the values we want to estimate for each segment. The MLP consists of several layers of neurons that can learn non-linear transformations of the input. The MLP can be trained to minimize the error between the prediction vector and the ground truth vector. The ground truth vector is the values we want to predict for each segment. Depending on what kind of challenge we are solving, we have different types of ground truth vectors and prediction vectors. For the VA challenge, we want to predict two values: valence and arousal. Valence measures how positive or negative an emotion is. Arousal measures how active or passive an emotion is. For the Expr challenge, we want to predict eight values: one for each basic expression (anger, disgust, fear, happiness, sadness, and surprise) plus neutral and other expressions. For the AU challenge, we want to predict twelve values: one for each action unit (AU1, AU2, AU4, AU6, AU7, AU10, AU12, 
AU15, AU23, AU24, AU25, AU26).

%-------------------------------------------------------------------------
\subsection{Loss Functions}
VA challenge: We use the Concordance Correlation Coefficient (CCC)  between the predictions and the ground truth labels as the measure, which is defined as in Eq \ref{eq1}. It measures the correlation between two sequences $x$ and $y$ and ranges between -1 and 1, 
where -1 means perfect anti-correlation, 0 means no correlation, and 1 means perfect correlation. The loss is calculated as Eq \ref{eq2}.

\begin{equation}\label{eq1}
\begin{split}
CCC(x, y) &= \frac{2 * \operatorname{cov}(x, y)}{\sigma_x^2+\sigma_y^2+\left(\mu_x-\mu_y\right)^2} \\
\text { where } \operatorname{cov}(x, y)&=\sum\left(x-\mu_x\right) *\left(y-\mu_y\right)
\end{split}
\end{equation}

\begin{equation}\label{eq2}
\mathcal{L}_{\text {VA }}=1-CCC
\end{equation}

Expr challenge: We use the cross-entropy loss as the loss function, which is defined as in Eq \ref{eq3}.
\begin{equation}\label{eq3}
\mathcal{L}_{\text {Expr }} = - \frac {1} {N}\sum_ {i} \sum_ {c=1}^My_ {ic}\log (p_ {ic})
\end{equation}

where $y_{ic}$ is a binary indicator (0 or 1) if class $c$ is the correct classification for observation $i$.
$p_{ic}$ is the predicted probability of observation $i$ being in class $c$, 
$M$ is the number of classes.
The multiclass cross entropy loss function measures how well a model predicts the true probabilities of each class for a given observation. It penalizes wrong predictions by taking the logarithm of the predicted probabilities. The lower the loss, the better the model.

AU challenge: We employ BCEWithLogitsLoss as the loss function, which integrates a sigmoid layer and binary cross-entropy,  which is defined as in Eq \ref{eq4}.

\begin{equation}\label{eq4}
\mathcal{L}_{\text {AU }} = - \frac {1} {N}\sum_ {i} [y_i\cdot log (\sigma (x_i)) + (1-y_i)\cdot log (1-\sigma (x_i))]
\end{equation}
where $N$ is the number of samples, $y_i$ is the target label for sample $i$, $x_i$ is the input logits for sample $i$, $\sigma$ is the sigmoid function
The advantage of using BCEWithLogitsLoss over BCELoss with sigmoid is that it can avoid numerical instability and improve performance.

\begin{table*}
  \centering
  \begin{tabular}{@{}lccccccc@{}}
    \toprule
    Task & Evaluation Metric  & Method & Fold 0 & Fold 1 & Fold 2 & Fold 3 & Fold 4 \\
    \midrule
    \multirow{2}*{Valence} & \multirow{4}*{CCC}  & Ours & 0.5385 & 0.6404 & 0.4926 & 0.5863 & 0.5403 \\
    ~                                 &                ~     & Baseline & 0.24     & - & - & - & -\\
        \cline{3-8}

    \multirow{2}*{Arousal} & ~  & Ours& 0.6224 & 0.5651 & 0.6015 & 0.6812 & 0.6342 \\
    ~ &~  & Baseline & 0.20     & - & - & - & -\\

    \midrule
     \multirow{2}*{Expr} & \multirow{2}*{ F1-score} & Ours & 0.4561 & 0.4478 & 0.4463 & 0.4583 & 0.4506\\
         ~ & ~ &Baseline &  0.23 & - & - & - & -\\

    \midrule
    \multirow{2}*{AU} & \multirow{2}*{F1-score}  & Ours &  0.5762 & 0.5566 & 0.5018 & 0.5556 & 0.5819\\
       ~ & ~  & Baseline & 0.39 & -& - & - & -\\

    \bottomrule
   \end{tabular}
   \caption{Results for the five folds of three tasks}
   \label{tab:results_fold}
\end{table*}

\section{Experiments and Results}
\label{sec:experiment}

%-------------------------------------------------------------------------
\subsection{Experiments Settings}
All models were trained on two Nvidia GeForce GTX 3090 GPUs with each having 24GB of memory. 
\subsubsection{MAE Pre-training}

We conducted an extensive pre-training of the MAE model on large-scale facial image datasets over 500 epochs, employing the AdamW optimizer. During this phase, we maintained a batch size of 1024 and set the learning rate to 0.0005. Subsequently, in the fine-tuning stage of MAE, we adjusted the batch size to 256 and lowered the learning rate to 0.0001, still leveraging the AdamW optimizer.

\subsubsection{Task Trainging}

We used the AdamW optimizer and cosine learning rate schedule with the first epoch warmup. The learning rate was set to $3e-5$, the weight decay to $1e-5$, the dropout probability to 0.3, and the batch size to 32.

Videos were split using a segment window of $w=300$ and a stride of $s=200$ for all three challenges. This meant we divided each video into segments of 300 frames with an overlap of 100 frames between consecutive segments. This approach helped capture the temporal dynamics of facial expressions and emotions.

%-------------------------------------------------------------------------
\subsection{Overall Results}
Table \ref{tab:results_fold} displays the experimental results of our proposed method on the validation set of the VA, Expr, and AU Challenge, where the Concordance Correlation Coefficient (CCC) is utilized as the evaluation metric for both valence and arousal prediction, and F1-score is used to evaluate the result of Expr and AU challenge. As demonstrated in the table, our proposed method outperforms the baseline significantly. These results show that our proposed approach using TCN and a Transformer-based model effectively integrates visual and audio information for improved accuracy in recognizing emotions on this dataset.

\section{Conclusion}\label{sec:conclusion}

Our proposed approach utilizes a combination of a Temporal Convolutional Network (TCN) and a Transformer-based model to integrate visual and audio information for improved accuracy in recognizing emotions. The TCN captures relationships at low-, intermediate-, and high-level time scales, while the Transformer mechanism merges audio and visual features. We conducted our experiment on the Aff-Wild2 dataset, which is a widely used benchmark dataset for emotion recognition. Our results show that our method significantly outperforms the baseline. 

{
    \small
    \bibliographystyle{ieeenat_fullname}
    \bibliography{main}
}

\end{document}